\pdfoutput=1

\documentclass[11pt]{article}

\usepackage{ACL2023}
\usepackage{soul}

\usepackage[final]{acl2023}

\usepackage{times}
\usepackage{latexsym}
\usepackage{booktabs}

\usepackage[T1]{fontenc}

\usepackage[utf8]{inputenc}

\usepackage{microtype}

\usepackage{inconsolata}

\usepackage{tabularx}
\usepackage{booktabs}  
\usepackage{amsmath, amssymb}
\usepackage{svg}
\usepackage{placeins}
\usepackage{subcaption}
\usepackage{hyperref}
\usepackage{float}

\usepackage{ulem}

\title{Mr. Snuffleupagus at SemEval-2025 Task 4: Unlearning Factual Knowledge from LLMs Using Adaptive RMU}

\author{Arjun Dosajh\thanks{\hspace{0.5em}Equal contribution.} \\
  IIIT Hyderabad \\
  \normalsize \href{mailto:arjun.dosajh@research.iiit.ac.in}{\texttt{arjun.dosajh@research.iiit.ac.in}} \\\And
  Mihika Sanghi\footnotemark[1] \\
  IIIT Hyderabad \\
  \normalsize \href{mailto:mihika.sanghi@research.iiit.ac.in}{\texttt{mihika.sanghi@research.iiit.ac.in}} \\
}

\begin{document}
\maketitle

\begin{abstract}
Large Language Models (LLMs) have demonstrated remarkable capabilities in natural language understanding and generation. However, their tendency to memorize training data raises concerns regarding privacy, copyright compliance, and security, particularly in cases involving Personally Identifiable Information (PII). Effective machine unlearning techniques are essential to mitigate these risks, yet existing methods remain underdeveloped for LLMs due to their open-ended output space. In this work, we apply the Adaptive Representation Misdirection Unlearning (RMU) technique to unlearn sensitive information from LLMs. Through extensive experiments, we analyze the effects of unlearning across different decoder layers to determine the most effective regions for sensitive information removal. Our technique ranked 4th on the official leaderboard of both 1B parameter and 7B parameter models.
\end{abstract}

\section{Introduction}
In the realm of large language models (LLMs), unlearning is particularly challenging due to the highly distributed nature of knowledge storage across model parameters. Unlike Computer Vision or Graph Neural Networks \cite{Chundawat_2023, kolipaka2024cognacshotforgetbad}, where feature representations tend to be more localized, LLMs encode knowledge in an interwoven manner, making targeted removal complex \cite{10.5555/3600270.3601532}. The need for effective LLM unlearning arises from concerns surrounding data privacy, bias mitigation, and compliance with regulations such as GDPR's `right to be forgotten'. If an LLM generates sensitive or misleading information, it is crucial to develop mechanisms that remove such knowledge without degrading overall performance \cite{eldan2023whosharrypotterapproximate}.

An approach to tackling unlearning in LLMs involves model-editing, a technique closely tied to mechanistic interpretability. Model-editing methods aim to modify the internal representations of an LLM to suppress or alter specific outputs while maintaining overall fluency and coherence. Previous work has explored various approaches, such as neuron pruning \cite{frankle2018the} and gradient-based forgetting \cite{yao-etal-2023-editing, goodfellow2015empiricalinvestigationcatastrophicforgetting}, to achieve unlearning without full retraining. Unlearning techniques for LLMs often lead to catastrophic forgetting, compromising general performance \cite{luo2025empiricalstudycatastrophicforgetting, zhang2024negativepreferenceoptimizationcatastrophic, Kemker_McClure_Abitino_Hayes_Kanan_2018, ROBINS01061995}. Understanding how and where knowledge is stored in an LLM is crucial for designing effective unlearning strategies. Research has shown that different decoder layers in transformer-based architectures capture different types of information.

The task is outlined in the task description paper \cite{ramakrishna2025semevaltask4}. There are separate leaderboards for the 1B and 7B parameter models.

Participating in this task provided valuable insights into both the strengths and limitations of our system. Quantitatively, our approach achieved competitive results, ranking 4th among participating teams on a metric designed specifically for this task. Additionally, we analyze how factual information is distributed across different decoder layers in large language models (LLMs). Through our experiments, we demonstrate that unlearning factual information from middle-later layers is particularly effective.

We have released the code for our system on GitHub\footnote{\href{https://github.com/ArjunDosajh/Mr.-Snuffleupagus-SemEval-2025-Task-4}{https://github.com/ArjunDosajh/Mr.Snuffleupagus-SemEval-2025-Task-4}},  
facilitating transparency and reproducibility in our approach.

\section{Related works}
Effective unlearning techniques focus on modifying model representations to diminish the influence of specific data while preserving overall performance. These methods often involve fine-tuning strategies that steer internal representations away from unwanted content, ensuring that the model forgets particular information without compromising its general capabilities.

One prominent approach is Representation Misdirection Unlearning (RMU) \cite{li2024wmdpbenchmarkmeasuringreducing}, which directs the model's intermediate representations of data intended for unlearning toward a predetermined random vector. This technique effectively reduces the model's performance on tasks related to the forgotten content while maintaining its proficiency in other domains. RMU has been demonstrated to lower the model's knowledge of the Weapons of Mass Destruction Proxy (WMDP) dataset, indicating its potential in reducing malicious use of LLMs \cite{li2024wmdpbenchmarkmeasuringreducing}. 

Building upon RMU, Adaptive RMU introduces a dynamic adjustment mechanism for the steering coefficient, which influences the alignment of forget-sample representations with the random direction \cite{huutien2025effectssteeringlatentrepresentation}. This adaptive approach enhances unlearning effectiveness across various network layers, addressing limitations observed when RMU is applied to middle and later layers of LLMs. Adaptive RMU not only improves unlearning performance but also maintains robustness against adversarial attacks, ensuring that the model does not inadvertently relearn the forgotten content. 

Another innovative technique involves the use of Sparse Autoencoders (SAEs) to remove specific knowledge from LLMs \cite{farrell2024applyingsparseautoencodersunlearn}. By training SAEs to capture and subsequently eliminate features associated with the content to be unlearned, this method effectively reduces the model's ability to recall unwanted information. Studies have shown that applying SAEs can unlearn subsets of sensitive data with minimal impact on the model's performance in other areas, offering a targeted and efficient unlearning strategy.

\section{Background}

\subsection{Task}
The challenge spans three subtasks evaluating unlearning: long-form synthetic creative texts across genres; short-form synthetic biographies containing PII (such as fake names, phone numbers, SSNs, email addresses, and home addresses); and real documents drawn from the model’s training data. Each subtask involves both sentence-completion and question-answering evaluations. In both cases, the dataset is divided into a retain set (which should be preserved) and a forget set (which should be unlearned). The goal is for the model to behave as if trained solely on the retain set, excluding the forget set, thereby mimicking an ideal unlearning scenario \cite{ramakrishna2025semevaltask4}.

\subsection{Dataset}

The dataset quantifies unlearning performance for each subtask and evaluation type, with separate retain and forget sets whose sizes are summarized in Table~\ref{tab:dataset_splits} \cite{ramakrishna2025lume}.

\begin{table}[h]
    \centering
    \begin{tabular}{ccc}
        \toprule
        \textbf{Subtask} & \textbf{Forget Set Size} & \textbf{Retain Set Size} \\
        \midrule
        1 & 214 & 260 \\
        2 & 780 & 762 \\
        3 & 372 & 392 \\
        \bottomrule
    \end{tabular}
    \caption{Dataset sizes for retain and forget sets across all subtasks.}
    \label{tab:dataset_splits}
\end{table}

\subsection{Objective}

The challenge aims to develop and evaluate effective unlearning techniques for large language models while preserving overall model performance. Submissions are assessed based on their ability to forget specified information while retaining non‑targeted knowledge.

Performance is measured through three metrics:

\begin{itemize}
  \item \textbf{Task‑Specific Regurgitation Rates:} This combines ROUGE‑L for sentence completion and exact‑match for question answering, evaluated on both the \emph{retain} and \emph{forget} sets. Forget‑set scores are inverted as \(1 - \text{value}\), and the 12 resulting scores are aggregated via the harmonic mean.
  
  \item \textbf{Membership Inference Attack (MIA) Score:} Calculated as
  \[
    1 - \bigl|\text{mia\_loss\_auc\_score} - 0.5\bigr| \times 2,
  \]
  this metric assesses privacy leakage following ~\cite{mia7958568}.
  
  \item \textbf{MMLU Benchmark Performance:} Accuracy on the 57‑subject multiple‑choice MMLU suite measures language understanding and reasoning~\cite{conf/iclr/HendrycksBBZMSS21}.
\end{itemize}

The final score is the arithmetic mean of these three metrics.

\subsection{Model}
For our experiments, we use fine-tuned versions of the \texttt{OLMo-7B-0724-Instruct-hf} and \texttt{OLMo-1B-0724-hf} models. OLMo (Open Language Model) is an open-source language model developed to facilitate research in language modeling and knowledge retention \cite{olmo_groeneveld2024olmoacceleratingsciencelanguage}. Both models follow a decoder-only transformer architecture, similar to traditional autoregressive language models. The task organizers provided the fine-tuned models.

The \texttt{OLMo-7B} model consists of \textbf{32 decoder layers} with a hidden size of 4096 and 32 attention heads. The smaller \texttt{OLMo-1B} model has \textbf{16 decoder layers}, a hidden size of 2048, and 16 attention heads. These models have been specifically fine-tuned to memorize documents from all three tasks, making them a suitable testbed for evaluating unlearning methods.

\begin{figure*}[!ht]  
    \centering
    \includegraphics[width=0.9\textwidth]{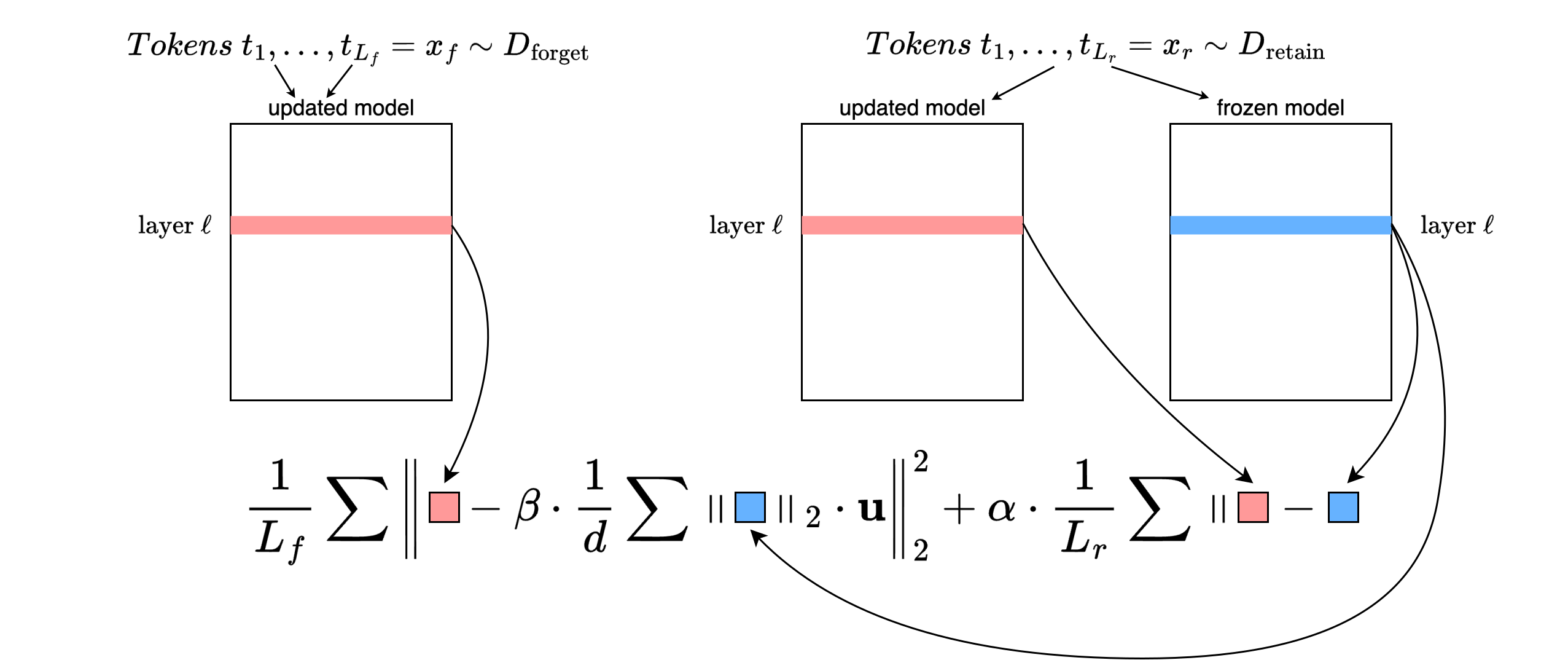}
    \caption{RMU performs machine unlearning by optimizing a two-part loss: a forget term, which misdirects the model's activations for forget-set inputs, and a retain term, which aligns the model's activations for retain-set inputs. The random vector $\boldsymbol{u}$ is adaptively scaled using the frozen model's activations, $\boldsymbol{d}$ denotes the embedding dimension of the model \cite{li2024wmdpbenchmarkmeasuringreducing, huutien2025effectssteeringlatentrepresentation}}
    \label{fig:adaptive_rmu}
\end{figure*}

\section{System Overview}
\subsection{RMU}

Representation Misdirection Unlearning (RMU) is a technique designed to selectively unlearn hazardous knowledge from a large language model (LLM) while preserving its general capabilities. It employs a two-part loss function: a \textit{forget loss} that degrades harmful representations and a \textit{retain loss} that ensures minimal disruption to benign knowledge.

The forget loss modifies model activations at a specific layer $\ell$ by increasing their norm in a fixed random direction. Given:
\begin{itemize}
    \setlength{\itemsep}{0pt}
    \item $M_u(\cdot)$: hidden states of the unlearned model
    \item $M_f(\cdot)$: hidden states of the original frozen model
    \item $\mathbf{u}$: a fixed random unit vector sampled from $[0,1]$
    \item $\mathcal{D}_f$: forget dataset
\end{itemize}
The forget loss is defined as:
\begin{equation}
    \mathcal{L}_{\text{forget}} = \mathbb{E}_{x_f \sim \mathcal{D}_f} \left[ \frac{1}{L_f} \sum_{t \in x_f} \left\| M_u(t) - c \cdot \mathbf{u} \right\|^2 \right],
\end{equation}
where $L_f$ is the number of tokens in $x_f$, and $c$ is a scaling hyperparameter.

To prevent excessive forgetting, RMU introduces a retain loss that regularizes the unlearned model activations to remain close to those of the frozen model on the retain dataset $\mathcal{D}_r$:
\begin{equation}
    \mathcal{L}_{\text{retain}} = \mathbb{E}_{x_r \sim \mathcal{D}_r} \left[ \frac{1}{L_r} \sum_{t \in x_r} \left\| M_u(t) - M_f(t) \right\|^2 \right],
\end{equation}
where $L_r$ is the number of tokens in $x_r$.

The final loss function is a weighted combination of the forget and retain losses:
\begin{equation}
    \mathcal{L} = \mathcal{L}_{\text{forget}} + \alpha \cdot \mathcal{L}_{\text{retain}},
\end{equation}
Where $\alpha$ controls the trade-off between forgetting and retention. RMU updates model weights iteratively, focusing on layers $\ell - 2, \ell - 1$, and $\ell$ to improve efficiency.

In the original RMU implementation, the authors use an external dataset, such as WikiText, to preserve the model's general capabilities. Instead, we adapt the retain dataset to match our task-specific retain set and demonstrate that RMU remains effective.

\subsection{Adaptive RMU}

Building on RMU, Adaptive RMU introduces a modified forget loss by scaling the random unit vector $\mathbf{u}$ with an \textit{adaptive scaling coefficient} $\beta \| h^{(l)}_{\theta_{\text{frozen}}}(x_F) \|$. Here, $\beta \in \mathbb{R}$ is a scaling factor, and $\| h^{(l)}_{\theta_{\text{frozen}}}(x_F) \|$ is the $\ell_2$-norm of the forget sample $x_F$ in the frozen model $f_{\theta_{\text{frozen}}}$. This ensures that the magnitude of the perturbation adapts to the norm of the activation, leading to a more stable unlearning process. The total loss in Adaptive RMU is given by:

\begin{small}
\begin{align}
\mathcal{L}^{\text{adaptive}} = \mathbb{E}_{x_f \sim \mathcal{D}_{\text{f}}} 
\left[ \frac{1}{L_f} \sum_{t \in x_f} \Big\| M_{\text{u}}(t) - \beta \| M_{\text{f}}(t) \| \mathbf{u} \Big\|_2^2 \right] \nonumber \\[1ex]
+ \alpha \mathbb{E}_{x_r \sim \mathcal{D}_{\text{r}}} 
\left[ \frac{1}{L_r} \sum_{t \in x_r} \Big\| M_{\text{u}}(t) - M_{\text{f}}(t) \Big\|_2^2 \right]
\end{align}
\end{small}

Figure~\ref{fig:adaptive_rmu} illustrates the construction of the loss function with adaptive scaling. This adaptation makes the forgetting process proportional to the activation strength of the original model while ensuring that general capabilities are preserved through the retain loss.

\noindent \newline To account for the varying dataset sizes across different splits (creative documents, sensitive content, and real documents), we use a randomized sampling approach instead of alternating uniformly between them. In each training step, a sample from both the forget and retain datasets is selected, tokenized, and processed by the model. To guide unlearning, each forget sample is assigned a control vector, which is scaled using a predefined steering coefficient. The unlearning loss is computed as the MSE between the model’s activations and the control vector, with an adaptive coefficient that dynamically scales based on the activation norms.

Due to computational constraints, our experiments were conducted exclusively on the 1B parameter model. However, our approach demonstrates competitive performance on the 7B parameter model as well.

\begin{table*}[!ht]
\centering
\begin{tabular}{ccccc}
\hline
\textbf{Decoder Layers} & \textbf{Task Aggregate} & \textbf{MIA} & \textbf{MMLU} & \textbf{Final score} \\
\hline
0,1,2 & 0.547 & 0.062 & 0.244 & 0.284 \\
1,2,3 & 0.542 & 0.081 & 0.249 & 0.291 \\
2,3,4 & 0.355 & 0.401 & 0.250 & 0.336 \\
3,4,5 & 0.433 & 0.490 & 0.254 & 0.392 \\
4,5,6 & 0.508 & 0.355 & 0.229 & 0.364 \\
5,6,7 & \textbf{0.637} & 0.357 & 0.262 & 0.419 \\
6,7,8 & 0.597 & 0.416 & 0.250 & 0.421 \\
7,8,9 & 0.616 & 0.332 & 0.245 & 0.398 \\
8,9,10 & 0.631 & 0.362 & \textbf{0.265} & 0.419 \\
9,10,11 & 0.574 & 0.471 & 0.264 & 0.437 \\
10,11,12 & 0.282 & 0.279 & 0.243 & 0.268 \\
11,12,13 & 0.582 & 0.489 & 0.254 & 0.442 \\
12,13,14 & 0.565 & \textbf{0.835} & 0.261 & \textbf{0.554} \\
13,14,15 & 0.538 & 0.747 & 0.258 & 0.515 \\
\hline
\end{tabular}
\caption{\label{results_table}
Performance of different layer combinations on task aggregate, MIA score, and MMLU score for OLMo-1B.
}
\end{table*}

\section{Experiments}
\subsection{Preprocessing}
Our preprocessing pipeline involves tokenizing the text using the allenai/OLMo-1B-0724-hf tokenizer, which has a vocabulary size of 50,280. This tokenizer includes specialized tokens for personally identifiable information (PII), such as email addresses and Social Security Numbers (SSNs), ensuring a structured representation of such entities within the model.

\subsection{Hyperparameter Tuning}
Since the adaptive RMU approach dynamically adjusts the scaling coefficient, we primarily focus on selecting the layers for unlearning. Specifically, we experiment with all possible combinations of three consecutive layers, ranging from (0,1,2) to (13,14,15). This allows us to identify the most effective layer range for minimizing interference with retained knowledge while ensuring effective unlearning.

\section{Results}

We evaluated the performance of our unlearned model using the evaluation metric described in Section 3.3. Table \ref{baselines} compares the performance of our approach with some baseline methods. The results of our experiments, including the task aggregate score, membership inference attack (MIA) score, and MMLU score across different layer combinations, are presented in Table \ref{results_table}. We find that ideal layers for unlearning with adaptive RMU are 12,13,14 for the 1B parameter model and 24,25,26 for the 7B parameter model. We conducted our experiments using four NVIDIA RTX 3090 GPUs, each equipped with 24GB of VRAM. Our approach ranked 4th among all competing teams on both the 1B parameter model leaderboard and the 7B parameter model leaderboard.

\begin{table*}
\centering
\begin{tabular}{ccccc}
\hline
\textbf{Method} & \textbf{Task Aggregate} & \textbf{MIA} & \textbf{MMLU} & \textbf{Final score} \\
\hline
\st{Gradient Ascent}	& \st{0} & \st{0.912} & \st{0.269} & \st{0.394} \\
\st{Gradient Difference}	& \st{0} & \st{0.382} & \st{0.348} & \st{0.243} \\
\st{KL Minimization} & \st{0} & \st{0.916}	& \st{0.269} & \st{0.395} \\
Negative Preference Optimization & 0.021 & 0.080 & 0.463 & 0.188 \\
Adaptive RMU & 0.387 & 0.872 & 0.485 & 0.376 \\

\hline
\end{tabular}
\caption{\label{baselines}
Comparison with baseline methods for OLMo-7B. Some methods have been striked out because MMLU score is below the minimum threshold of 0.371.
}
\end{table*}

\section{Conclusion}
In this work, we demonstrate that applying the adaptive RMU (Rank-One Model Update) technique to later decoder layers is an effective strategy for unlearning factual information from large language models (LLMs). A key advantage of our method is its minimal hyperparameter tuning requirement, making it easily adaptable to different LLM architectures.

Our technique achieved 4th place on the official leaderboard for both 1B and 7B parameter models, highlighting its competitiveness among various unlearning approaches. While the original RMU implementation demonstrated that unlearning from earlier layers effectively removes hazardous knowledge—such as information related to cybersecurity threats and bioweapons, which require deep contextual understanding—we extend this research by showing that factual knowledge (e.g., phone numbers, Social Security Numbers, and addresses) is more effectively unlearned from later decoder layers.

Despite these promising results, unlearning sensitive content from LLMs remains an open challenge, with several promising directions for future research. For instance, further exploration is needed to assess the trade-offs between unlearning effectiveness and generalization, particularly when removing factual knowledge versus conceptual reasoning.

Overall, our work contributes to the broader fields of model editing and mechanistic interpretability, providing valuable insights into the layerwise dynamics of unlearning and paving the way for more robust and efficient strategies in the future.

\bibliography{custom}
\bibliographystyle{acl_natbib}

\appendix
\section{Appendix: Analysis of layer combinations for unlearning}
\begin{figure*}[!ht]
    \centering
    \begin{subfigure}[b]{0.48\textwidth}
        \includegraphics[width=\textwidth]{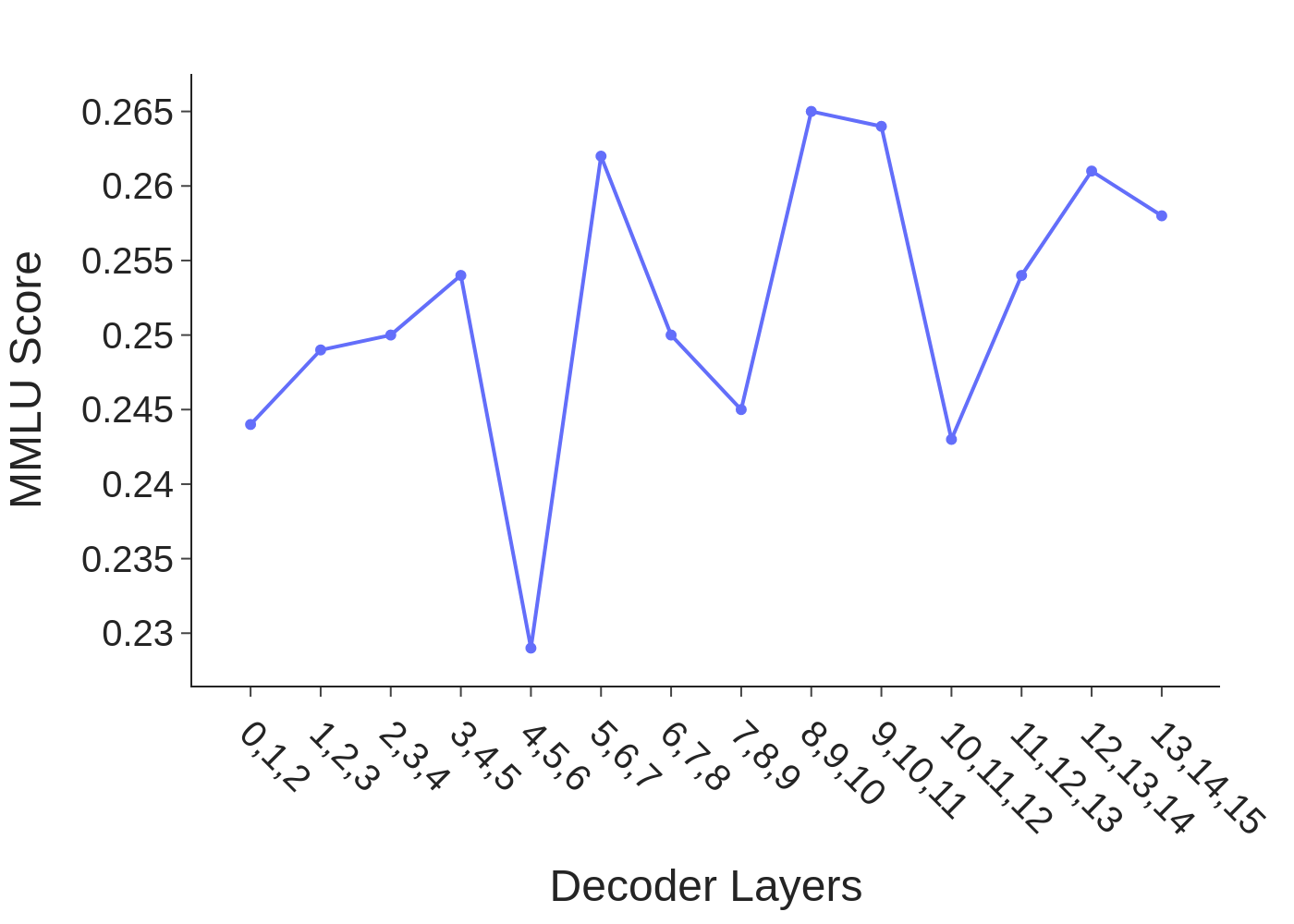}
        \caption{MMLU Score}
        \label{fig:2a}
    \end{subfigure}
    \hfill
    \begin{subfigure}[b]{0.48\textwidth}
        \includegraphics[width=\textwidth]{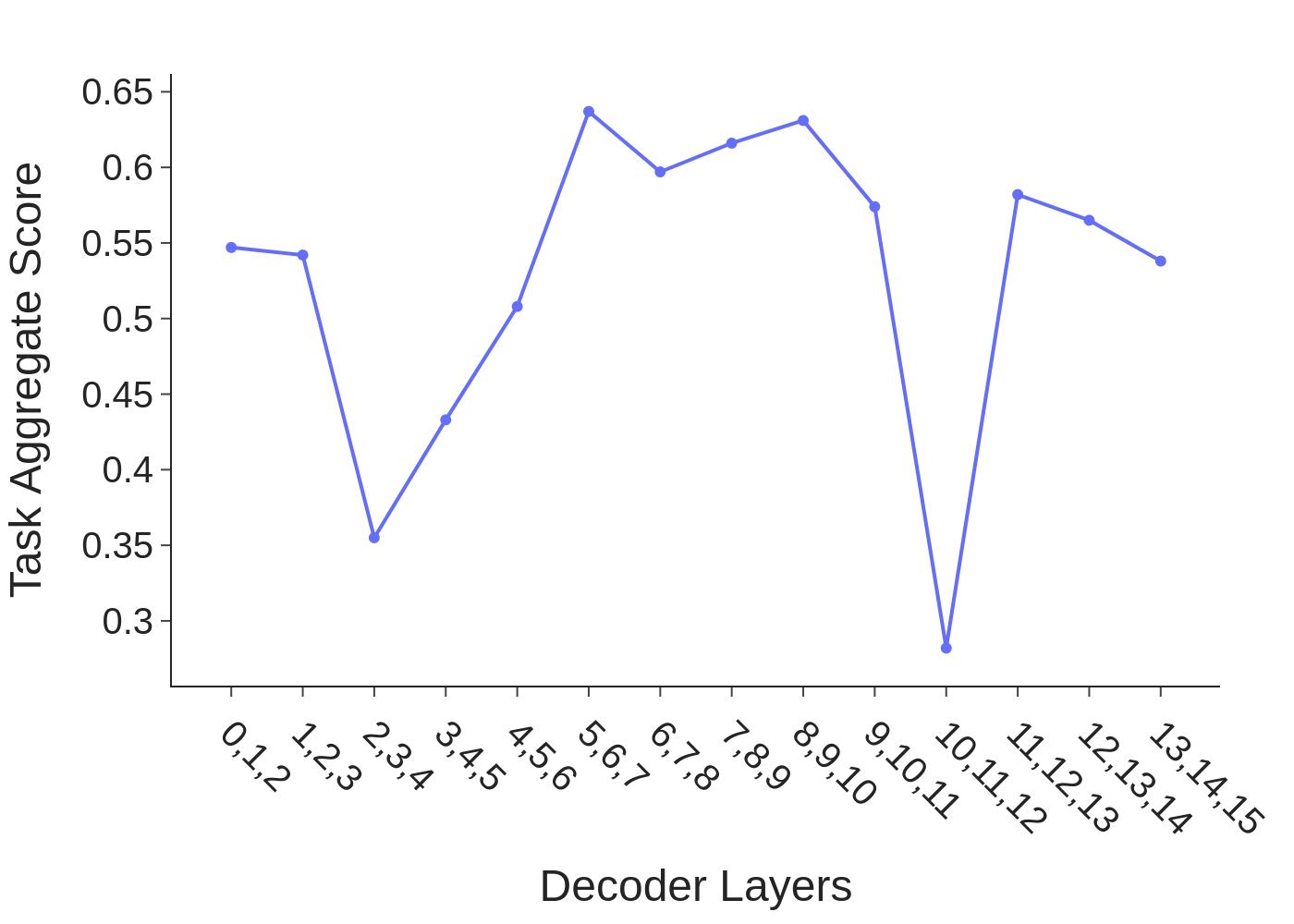}
        \caption{Task Aggregate Score}
        \label{fig:2b}
    \end{subfigure}
    \vspace{0.5cm}
    \begin{subfigure}[b]{0.48\textwidth}
        \includegraphics[width=\textwidth]{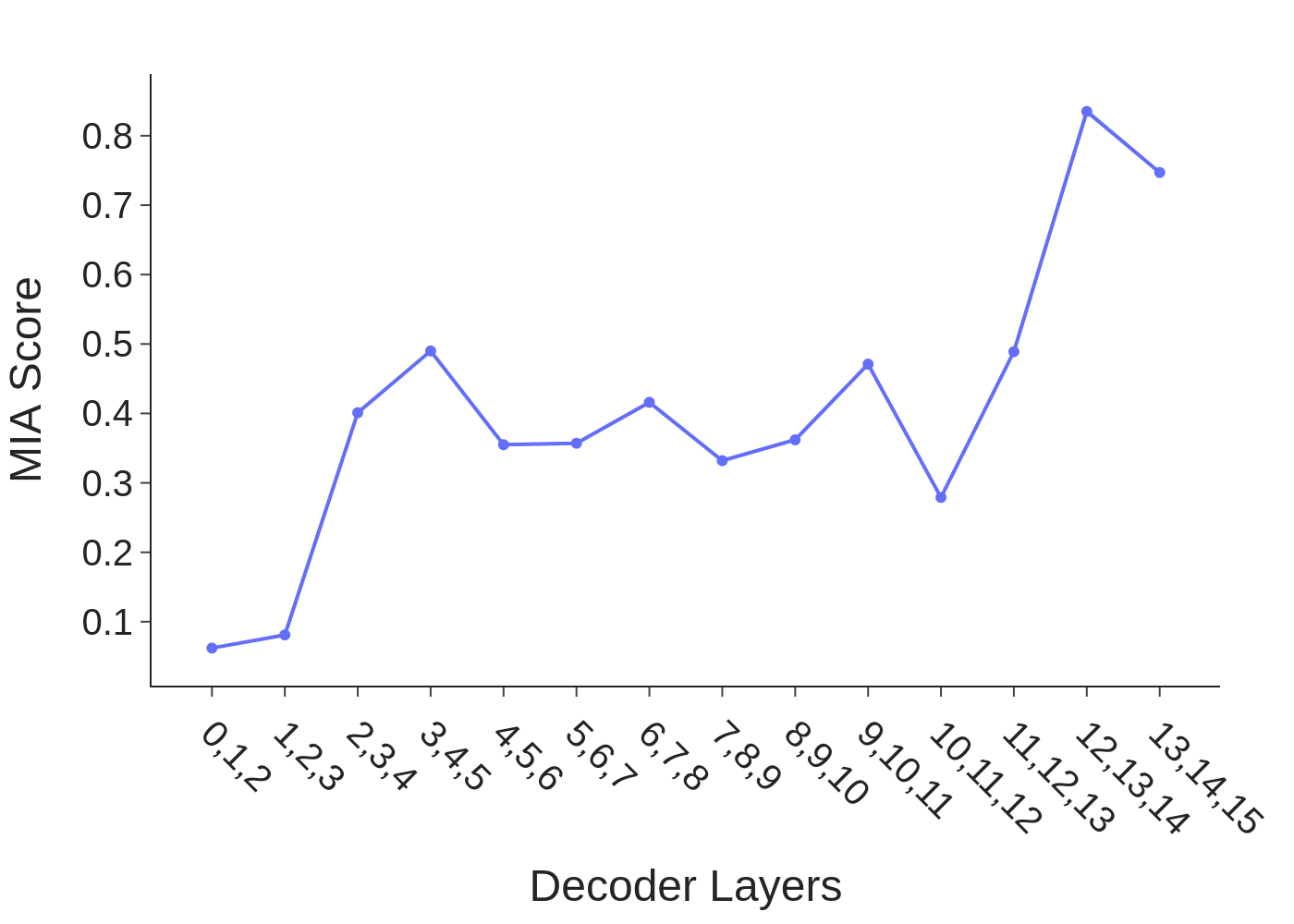}
        \caption{MIA Score}
        \label{fig:2c}
    \end{subfigure}
    \hfill
    \begin{subfigure}[b]{0.48\textwidth}
        \includegraphics[width=\textwidth]{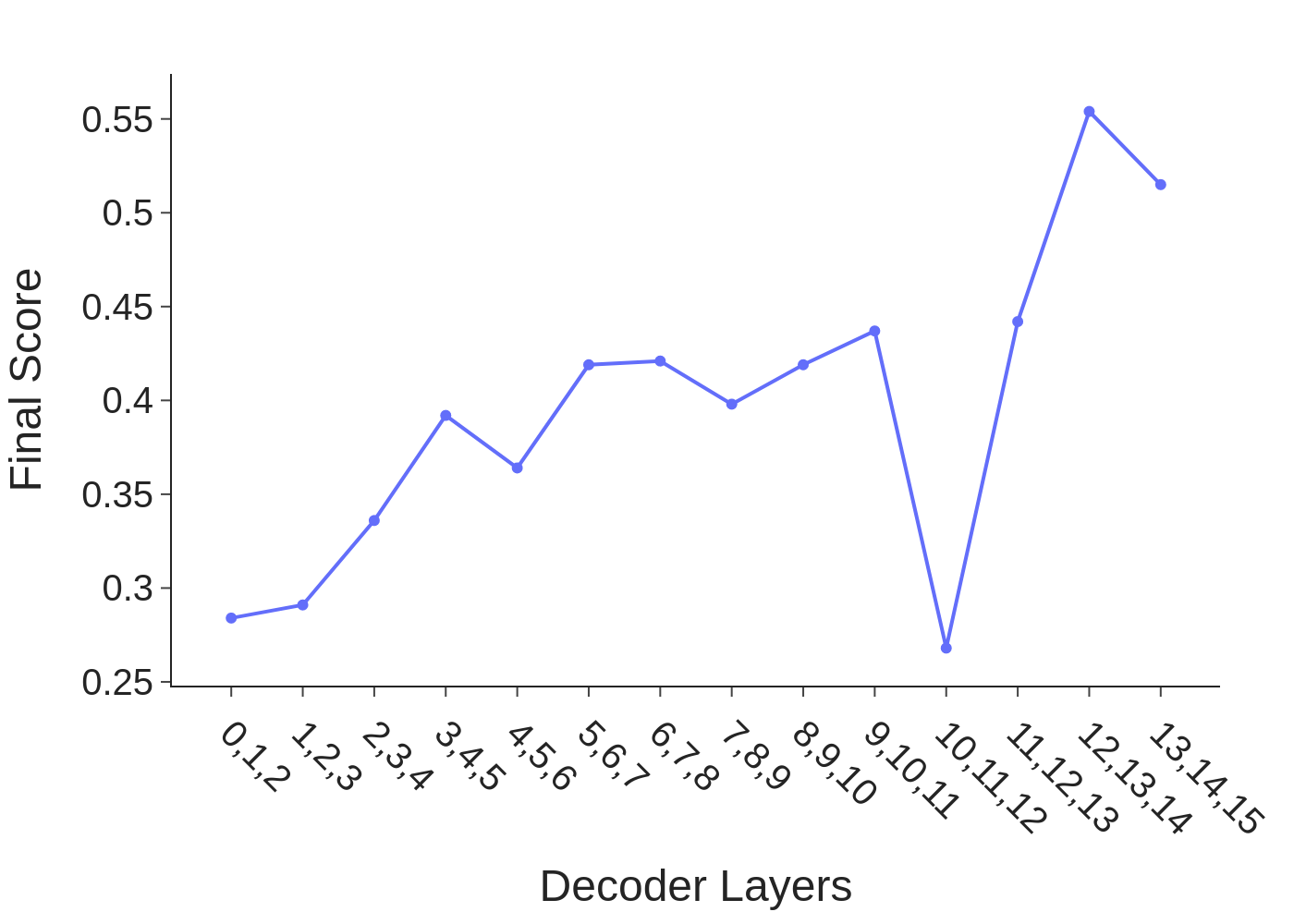}
        \caption{Final Score}
        \label{fig:2d}
    \end{subfigure}
    \caption{Observed trends of the MMLU Score (a), Task Aggregate Score (b), MIA Score (c), and Final Score (d) vs. unlearned decoder layers.}
    \label{fig:combined_plots}
\end{figure*}

Figure~\ref{fig:combined_plots}(a),~\ref{fig:combined_plots}(b) show that unlearning from the middle layers achieves a balance between knowledge retention (high MMLU score) and unlearning performance (high Task Aggregate score), but it remains more susceptible to MIA (Figure~\ref{fig:combined_plots}(c)). In contrast, later layers exhibit significantly higher robustness to MIA. The substantial improvement in MIA robustness outweighs the relatively smaller decline in knowledge retention and unlearning performance, making later layers the more effective choice for unlearning (Figure~\ref{fig:combined_plots}(d)).

\end{document}